\documentclass[11pt]{article}
\usepackage{geometry}
\usepackage{fancyhdr}
\usepackage{amsfonts,amsmath}
\usepackage{algorithm}
\usepackage[noend]{algorithmic}
\usepackage{graphicx}

\newcommand {\N} {\ensuremath {\mathcal{N}}}
\newcommand {\IE} {\ensuremath {\mathbb{E}}}
\newcommand {\BEL} {\ensuremath {\mathrm{BEL}}}

\begin{document}

\begin{center}
\large\textbf{Rational Value~of~Information Estimation for Measurement~Selection} \\
\vspace{2em}
\normalsize\textbf{David Tolpin} \\
\normalsize\textbf{Computer Science Dept., Ben-Gurion University, 84105 Beer-Sheva, Israel}
\vspace{1em} \\
\normalsize\textbf{Solomon Eyal Shimony} \\
\normalsize\textbf{Computer Science Dept., Ben-Gurion University, 84105 Beer-Sheva, Israel}
\end{center}

\vspace{1em}
\paragraph{ABSTRACT.}
Computing value of information (VOI) is a crucial task in various
aspects of decision-making under uncertainty, such as in
meta-reasoning for search; in selecting measurements to make, prior to
choosing a course of action; and in managing the exploration
vs. exploitation tradeoff. Since such applications typically require
numerous VOI computations during a single run, it is essential that
VOI be computed efficiently.  We examine the issue of anytime
estimation of VOI, as frequently it suffices to get a crude estimate
of the VOI, thus saving considerable computational resources. As a
case study, we examine VOI estimation in the measurement selection
problem.  Empirical evaluation of the proposed scheme in this domain
shows that computational resources can indeed be significantly
reduced, at little cost in expected rewards achieved in the overall
decision problem.
\vspace{1em}

\section{\normalsize INTRODUCTION}

Problems of decision-making under uncertainty frequently contain
cases where information can be obtained using some costly actions,
called measurement actions. In order to act rationally in the
decision-theoretic sense, measurement plans are typically optimized based on
some form of value of information (VOI). Computing VOI can
also be computationally intensive. Since frequently an exact
VOI is not needed in order to proceed (e.g. it is sufficient to determine
that the VOI of a certain measurement is much lower than that of
another measurement, at a certain point in time), significant computational
resources can be saved by controlling the resources used for estimating the
VOI. This paper examines this tradeoff via a case study of
measurement selection.

\sloppy
In general, computation of value of information (VOI), even under the
commonly used simplifying myopic assumption, involves multidimensional
integration of a general function \cite{Russell.right}. For some
problems, the integral can be computed efficiently
\cite{Russell.gametree}; but when the utility function is
computationally intensive or when a non-myopic estimate is used, the
time required to compute the value of information can be significant
\cite{Heckerman.nonmyopic} \cite{BilgicGetoor.voila} and must be taken
into account while computing the net value of information.  This paper
presents and analyzes an extension of the known greedy
algorithm that decides when to recompute VOI of
each of the measurements based on the principles of limited
rationality \cite{Russell.right}.

\sloppy
Although it may be possible to use this idea in more general settings,
this paper mainly examines on-line most informative measurement
selection \cite{Guestrin.submodular} \cite{BilgicGetoor.voila}, an
approach which is commonly used to solve problems of optimization
under uncertainty \cite{Rish.efficient} \cite{Krause.water}. Since this
approach assumes that the computation time required to select the most
informative measurement is negligible compared to the measurement
time\cite{Russell.right}, it is important in this setting to ascertain
that VOI estimation indeed does not consume excessive computational
resources.


\section{\normalsize THE MEASUREMENT SELECTION PROBLEM}
\label{sec:greedy}

As our case study, we examine the following optimization problem. Given:
\begin{itemize}
\item A set of $N_s$ items $S=\{s_1, s_2, \ldots, s_{N_s}\}$.
\item A set of $N_f$ item features $Z=\{ z_1, z_2, \ldots, z_{N_f}\}$. (Each feature
$z_i$ has a domain $\mathcal{D}(z_i)$.)
\item A joint distribution over the features of the items in $S$. That is,
a joint distribution over the random
variables $\{ z_1(s_1), z_2(z_1),\ldots , z_1(s_2), z_2(s_2),\ldots\} $.
\item A set of measurement types $M=\{(c, p)_k\:\vline\; k \in
  1..N_m\}$, with potentially different intrinsic measurement cost
  $c$ and observation distribution $p$, conditional on the true
  feature values, for each measurement type.
\item A utility function $u(\textbf{z})\colon\mathbb{R}^{N_f}\to
  \mathbb{R}$ on features. In the simplest case, there is just one real-valued
feature, acting as the item's utility value, and $u$ is simply the identity function.
\item A measurement budget $C$.
\end{itemize}
Find a policy of measurement decisions and a final selection
that maximize the expected net utility of the selection (the expected reward):
\begin{equation}
\label{eq:problem-reward}
\mbox{max:}R=u(\textbf{z}(s_\alpha))-\sum_{i=1}^{N_q}c_{k_i}\quad\mbox{s.t.:}\sum_{i=1}^{N_q}c_{k_i}\le C
\end{equation}
where $Q=\{(k_i,s_i)\:\vline\;i \in
1..N_q\}$ is the performed measurement sequence and $s_\alpha$ is the selected item. A next measurement is selected on-line, after the outcomes of all
preceding measurements are known.

The above selection problem is intractable, and is therefore commonly solved
approximately using a greedy heuristic algorithm. The greedy algorithm
selects a measurement $m_{j_{\max}}$ with the greatest net value of information
$V_{j_{\max}}$. The {\it net value of information} is the difference
between the intrinsic value of information and the measurement cost.
\begin{equation}
\label{eq:greedy-net-value}
V_j = \Lambda_j -c_{k_j}
\end{equation}
The {\it intrinsic value of information} $\Lambda_j$ is the expected
difference in the true utility of the finally selected item $s_\alpha$
after and before the measurement.
\begin{equation}
\label{eq:greedy-intrinsic-value}
\Lambda_j = \IE(\IE[u(\textbf{z}(z_{\alpha^j}))]-\IE[u(\textbf{z}(s_\alpha))])
\end{equation}
Exact computation of $\Lambda_j$ is intractable, and various estimates
are used, including the myopic estimate  \cite{Russell.right} and semi-myopic
schemes \cite{TolpinShimony.blinkered}.

The pseudocode for the algorithm is presented as Algorithm~\ref{alg:greedy}. 
\begin{algorithm}[t]
\caption{Greedy measurement selection}
\label{alg:greedy}
\begin{algorithmic}[1]
\STATE $budget \leftarrow C$
\STATE Initialize beliefs                             \label{alg:greedy-initialize-beliefs}
\LOOP                                                 \label{alg:greedy-main-loop-start}
  \FORALL {items $s_i$} 
    \STATE Compute $\IE(U_i)$
  \ENDFOR
  \FORALL {measurements $m_j$}                        \label{alg:greedy-voi-start}
    \IF {$c_j \le budget$}                            \label{alg:greedy-within-budget}
      \STATE Compute $V_j$
    \ELSE
      \STATE $V_j \leftarrow 0$
    \ENDIF
  \ENDFOR                                            \label{alg:greedy-voi-end}
  \STATE $j_{\max} \leftarrow \arg \max\limits_j V_j$  \label{alg:greedy-select}
  \IF {$V_{j_{\max}}>0$}                                \label{alg:greedy-positive-value}
    \STATE Perform measurement $m_{j_{\max}}$; Update beliefs; $budget \leftarrow  budget-c_{j_{\max}}$ \label{alg:greedy-measure} \label{alg:greedy-update-beliefs}
  \ELSE                       
    \STATE {\bf break}                              \label{alg:greedy-break}
  \ENDIF
\ENDLOOP                                            \label{alg:greedy-main-loop-end}
\STATE $\alpha \leftarrow \arg \max \limits_j \IE(U_i)$
\RETURN $s_\alpha$                                   \label{alg:greedy-return-alpha}
\end{algorithmic}
\end{algorithm}
At each step, the algorithm
recomputes the value of information estimate of every
measurement.
The assumptions behind the greedy algorithm are justified when the
cost of selecting a next measurement is negligible compared to the
measurement cost. However, optimization problems with hundreds and
thousands of items are common \cite{TolpinShimony.blinkered}; and even
if the value of information of a single measurement can be computed
efficiently \cite{Russell.gametree}, the cost of estimating the value
of information of all measurements becomes comparable to and outgrows
the cost of performing a measurement.

Recomputation of the value of information for every measurement is
often unnecessary, especially when using the "blinkered" scheme
\cite{TolpinShimony.blinkered}, a greedy algorithm which
attempts to also compute VOI for {\em sequences} of measurements of
the same type. When there are many different measurements, the value
of information of most measurements is unlikely to change abruptly due
to just one other measurement results. With an appropriate uncertainty
model, it can be shown that the VOI of only a few of the measurements
must be recomputed after each measurement, thus decreasing the
computation time and ensuring that the greedy algorithm exhibits a
more rational behavior w.r.t. computational resources.

\section{\normalsize RATIONAL COMPUTATION OF VALUE OF INFORMATION}
\label{sec:rational}

For the selective VOI recomputation, the belief
$\BEL(\Lambda_j)$ about the intrinsic value of information
of measurement $m_j$ is modeled by a normal distribution with
variance $\varsigma_j^2$:
\begin{equation}
\BEL(\Lambda_j)=\N(\Lambda_j, \varsigma_j^2)
\end{equation}
After a measurement is performed, and the beliefs about the item
features are updated
(line~\ref{alg:greedy-update-beliefs} of
Algorithm~\ref{alg:greedy}), the belief about $\Lambda_j$ becomes less
certain. Under the assumption that the influence of each measurement
on the value of information of other measurements is independent of
influence of any other measurement, the uncertainty is expressed by
adding Gaussian noise with variance $\tau^2$ to the belief:
\begin{equation}
\varsigma_j^2 \leftarrow \varsigma_j^2+\tau^2
\label{eq:varsigma}
\end{equation}
When $\Lambda_j$ of measurement $m_j$ is computed, $\BEL(\Lambda_j)$
becomes exact ($\varsigma_j^2 \leftarrow 0$). At the beginning of the
algorithm, the beliefs about the intrinsic value of information of
measurements are computed from the initial beliefs about item
features.

In the algorithm that recomputes the value of information selectively, the initial beliefs
about the intrinsic value of information are computed immediately after
line~\ref{alg:greedy-initialize-beliefs} in
Algorithm~\ref{alg:greedy}, and
lines~\ref{alg:greedy-voi-start}--\ref{alg:greedy-select} of
Algorithm~\ref{alg:greedy} are replaced by Algorithm~\ref{alg:rational}.
\begin{algorithm}[t]
\caption{Rational computation of the value of information}
\label{alg:rational}
\begin{algorithmic}[1]
  \FORALL {measurements $m_j$}
     \IF {$c_j \le budget$}
       \STATE $V_j\leftarrow \Lambda_j-c_j;\quad\varsigma_j \leftarrow \sqrt {{\varsigma_j}^2+\tau^2}$
     \ELSE
       \STATE $V_j \leftarrow 0;\quad\varsigma_j \leftarrow 0$
     \ENDIF
  \ENDFOR
  \LOOP
    \FORALL {measurements $m_k$}                            \label{alg:rational-voi-start}
       \IF {$c_k \le budget$}
         \STATE Compute $W_k$
       \ELSE 
         \STATE $W_k \leftarrow 0$
       \ENDIF
    \ENDFOR                                                 \label{alg:rational-voi-end}
    \STATE $k_{\max} \leftarrow \arg \max\limits_k W_k$       \label{alg:rational-select}
    \IF {$W_{k_{\max}} \le 0$}
      \STATE {\bf break}
    \ENDIF
    \STATE Compute $\Lambda_{k_{\max}};\quad 
           V_{k_{\max}}\leftarrow \Lambda_{k_{\max}}-c_{k_{\max}};\quad \
           \varsigma_{k_{\max}}
      \leftarrow 0$                                         \label{alg:rational-compute}
  \ENDLOOP
  \STATE $j_{\max} \leftarrow \arg\max_j V_j$
  \STATE Compute $\Lambda_{j_{\max}};\quad 
           V_{j_{\max}}\leftarrow \Lambda_{j_{\max}}-c_{j_{\max}};\quad \
           \varsigma_{j_{\max}}  
      \leftarrow 0$                                        \label{alg:rational-compute-chosen}
\end{algorithmic}
\end{algorithm}
While the number of iterations in 
lines~\ref{alg:rational-voi-start}--\ref{alg:rational-select}
of Algorithm~\ref{alg:rational} is the same as in
lines~\ref{alg:greedy-voi-start}--\ref{alg:greedy-voi-end}
of Algorithm~\ref{alg:greedy}, $W_k$ is efficiently computable,
and the subset of measurements for which the value of information is computed in
line~\ref{alg:rational-compute} of Algorithm~\ref{alg:rational} is
controlled by the computation cost $c_V$:
\begin{equation}
\label{eq:rational-w}
W_k=\frac {\varsigma_k} {\sqrt {2\pi}}
   e^{\left(-\frac {\left(V_\gamma-V_k\right)^2}
         {2\varsigma_k^2}\right)}
   -\left|V_\gamma-V_k\right|\Phi\left(-\frac {\left|V_\gamma-V_k\right|} {\varsigma_k} \right)-c_V
\end{equation}
where $V_\gamma$ is the highest value of information $V_\alpha$ if any but the highest
value of information is recomputed, and the next to highest value of information $V_\beta$ if the
highest value of information is recomputed; $\Phi(x)$ is the Gaussian cumulative
probability of $x$ for $\mu=0,\;\sigma=1$.

\subsection{Obtaining Uncertainty Parameters}
Uncertainty variance $\tau^2$ can be learned as a function of the
total cost of performed measurements, either off-line from earlier
runs on the same class of problems, or on-line. Learning $\tau^2(c)$ on-line from
earlier VOI recomputations  proved to be robust and easy to implement:
$\tau^2$ is initialized to $0$ and gradually updated with
each recomputation of the value of information.

\section{\normalsize EMPIRICAL EVALUATION}
\label{sec:empirical}

Experiments in this section compare performance of the algorithm that recomputes the value of
information selectively with the original algorithm in which the value of
information of every measurement is recomputed at every step.
Two of the problems evaluated in \cite{TolpinShimony.blinkered} are 
considered: {\it noisy Ackley function
maximization} and {\it SVM parameter search}.
For each of the optimization problems, plots of the number of VOI
recomputations, the reward, the intrinsic utility, and the total cost
of measurements are presented. The results are averaged for multiple
(100) runs of each experiment, such that the standard deviation of the
reward is $\approx 5\%$ of the mean reward. In the plots, the solid line
corresponds to the rationally recomputing algorithm, the dashed line
corresponds to the original algorithm, and the dotted line corresponds
to the algorithm that selects measurements randomly and performs the
same number of measurements as the rationally recomputing algorithm
for the given computation cost $c_V$. Since, as can be derived from
(\ref{eq:rational-w}), the computation time $T_r$ of the rationally
recomputing algorithm decreases with the logarithm of the computation
cost $c_V$, $T_r = \Theta(A-B\log c_V)$, the computation cost axis is
scaled logarithmically.

\subsection{The Ackley Function}

The Ackley function \cite{Ackley.function} is a popular optimization
benchmark. The two-argument form of the Ackley function is used in the
experiment; the function is defined by the expression (\ref{eq:emp-ackley}):
\begin{equation}
\label{eq:emp-ackley}
A(x,y)=20\cdot \exp\left(-0.2\sqrt { \frac {x^2+y^2} 2}\right)+\exp\left(\frac{cos(2\pi x)+cos(2\pi y)} 2\right)
\end{equation}
In the optimization problem, the utility function is
$u(z)=\tanh(2z)$, the measurements are normally distributed around the
true values with variance $\sigma_m^2=0.5$, and the measurement cost is
$0.01$. There are uniform dependencies with $\sigma_w^2=0.5$ in both
directions of the coordinate grid with a step of $0.2$ along each axis.
\begin{figure}[h]
\centering
\includegraphics[scale=0.63]{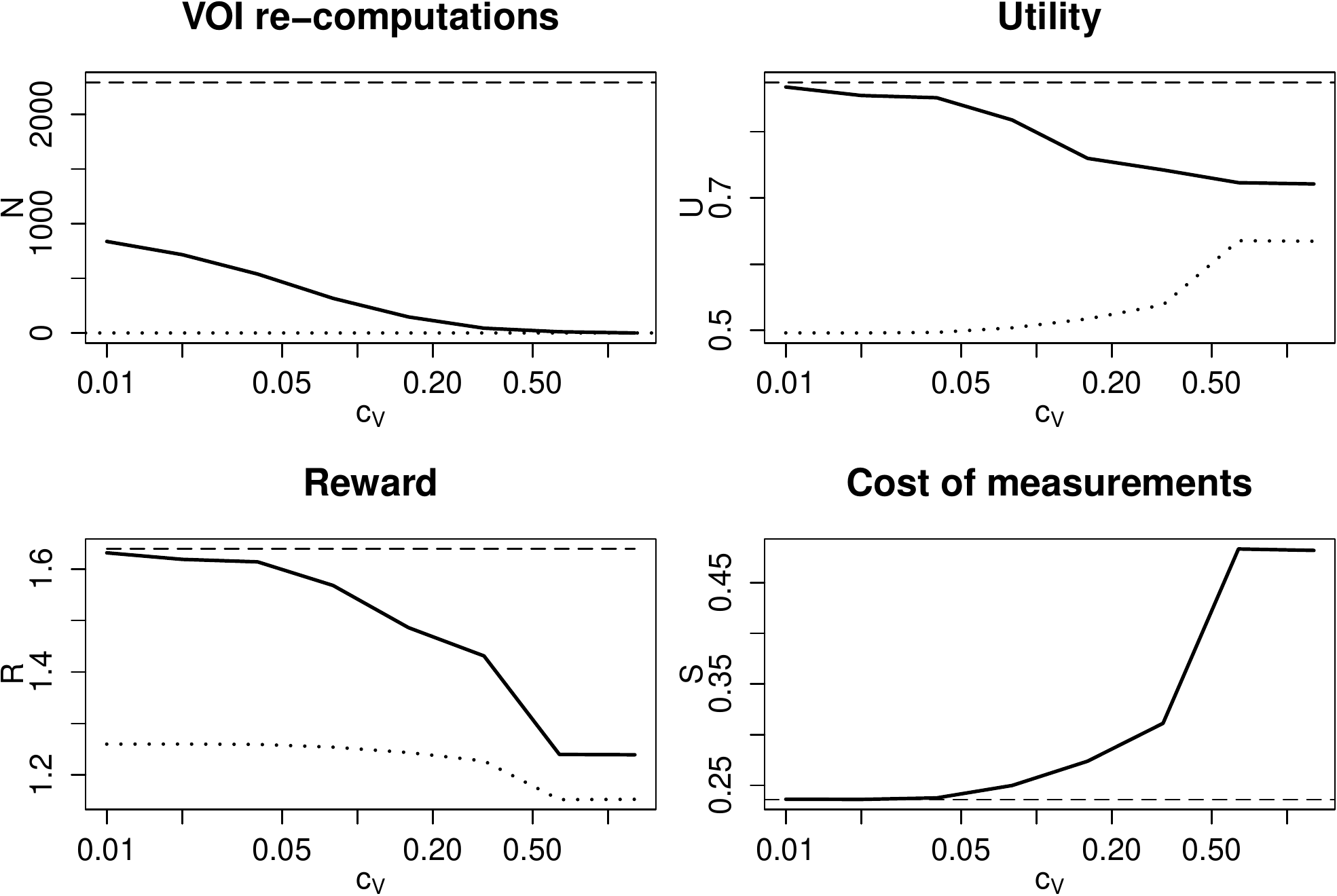}
\caption{The Ackley function, blinkered scheme.}
\label{fig:ackley-blinkered}
\end{figure}
The results for the blinkered scheme\cite{TolpinShimony.blinkered}
are presented in Figure~\ref{fig:ackley-blinkered}.

\subsection{SVM Parameter Search}

An SVM (Support Vector Machine) classifier based on the radial basis
function has two parameters: $C$ and $\gamma$.  A combination of $C$
and $\gamma$ with high expected classification accuracy should be
chosen, and an efficient algorithm for determining the optimal values
is not known. A trial for a combination of parameters determines
estimated accuracy of the classifier through cross-validation. The
{\sc svmguide2} \cite{Hsu.dataset} dataset is used for the case study.
The utility function is $u(z)=\tanh(4(z-0.5))$, the $\log C$ and $\log
\gamma$ axes are scaled for uniformity to ranges $[1..21]$ and there
are uniform dependencies along both axes with $\sigma_w^2=0.4$. The
measurements are normally distributed with variance $\sigma_m^2=0.25$
around the true values, and the measurement cost is $c_m=0.01$.
\begin{figure}[h]
\centering
\includegraphics[scale=0.63]{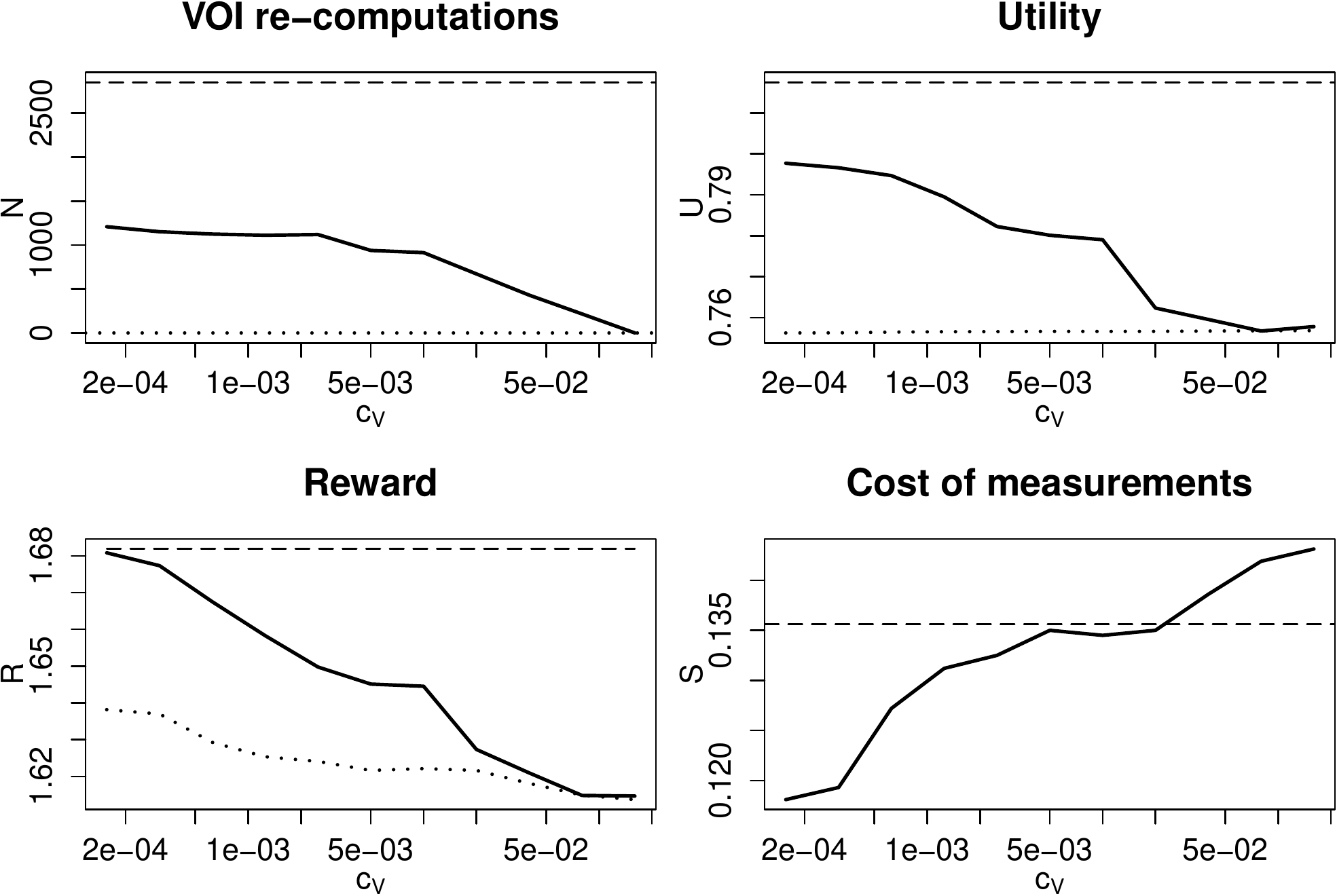}
\caption{SVM parameter search, myopic scheme.}
\label{fig:svmguide2}
\end{figure}
The results for the myopic scheme are presented in
Figure~\ref{fig:svmguide2}. 

\subsection{Discussion of Results}

In all experiments, a significant decrease in the computation time is
achieved with only slight degradation of the reward; performance of
the rationally recomputing algorithm decreases slowly with the
computation cost and exceeds performance of the algorithm that
makes random measurements even when VOI for only a small fraction of
measurements is recomputed at each step. Exact dependency of
performance of the rationally recomputing of algorithm on the
intensity of VOI recomputations varies among problems and depends
both on the problem properties and on the VOI estimate used in the
algorithm.

\section{\normalsize CONCLUSION}
\label{sec:conclusion}
The paper proposes an improvement to a widely used class of VOI-based
optimization algorithms. The improvement allows to decrease the computation time
while only slightly affecting the performance. The proposed algorithm
rationally reuses computations of VOI and recomputes VOI only for
measurements for which a change in VOI is likely to affect the choice
of the next measurement.

\section{\normalsize ACKNOWLEDGEMENTS}

The research is partially supported by the IMG4 Consortium under the MAGNET
program of the Israeli Ministry of Trade and Industry, by Israel
Science Foundation grant 305/09, by the Lynn and William Frankel
Center for Computer Sciences, and by the Paul Ivanier Center for
Robotics Research and Production Management.

\bibliographystyle{apalike}
\bibliography{refs}

\begin{thebibliography}{}

\bibitem[Ackley, 1987]{Ackley.function}
Ackley, D.~H. (1987).
\newblock {\em A connectionist machine for genetic hillclimbing}.
\newblock Kluwer Academic Publishers, Norwell, MA, USA.

\bibitem[Bilgic and Getoor, 2007]{BilgicGetoor.voila}
Bilgic, M. and Getoor, L. (2007).
\newblock Voila: Efficient feature-value acquisition for classification.
\newblock In {\em AAAI}, pages 1225--1230. AAAI Press.

\bibitem[Heckerman et~al., 1993]{Heckerman.nonmyopic}
Heckerman, D., Horvitz, E., and Middleton, B. (1993).
\newblock An approximate nonmyopic computation for value of information.
\newblock {\em IEEE Trans. Pattern Anal. Mach. Intell.}, 15(3):292--298.

\bibitem[Krause and Guestrin, 2007]{Guestrin.submodular}
Krause, A. and Guestrin, C. (2007).
\newblock Near-optimal observation selection using submodular functions.
\newblock In {\em AAAI}, pages 1650--1654.

\bibitem[Krause et~al., 2008]{Krause.water}
Krause, A., Leskovec, J., Guestrin, C., VanBriesen, J., and Faloutsos, C.
  (2008).
\newblock Efficient sensor placement optimization for securing large water
  distribution networks.
\newblock {\em Journal of Water Resources Planning and Management},
  134(6):516--526.
\newblock (Draft; full version available here).

\bibitem[Russell and Wefald, 1989]{Russell.gametree}
Russell, S.~J. and Wefald, E. (1989).
\newblock On optimal game-tree search using rational meta-reasoning.
\newblock In {\em IJCAI}, pages 334--340.

\bibitem[Russell and Wefald, 1991]{Russell.right}
Russell, S.~J. and Wefald, E. (1991).
\newblock {\em Do the right thing: studies in limited rationality}.
\newblock MIT Press, Cambridge, MA, USA.

\bibitem[Tolpin and Shimony, 2010]{TolpinShimony.blinkered}
Tolpin, D. and Shimony, S.~E. (2010).
\newblock Semi-myopic measurement selection for optimization under uncertainty.
\newblock Technical Report 10-01, Lynne and William Frankel Center for Computer
  Science at Ben Gurion University of the Negev, Israel.

\bibitem[wei Hsu et~al., 2003]{Hsu.dataset}
wei Hsu, C., chung Chang, C., and jen Lin, C. (2003).
\newblock A practical guide to support vector classification.
\newblock Technical report.

\bibitem[Zheng et~al., 2005]{Rish.efficient}
Zheng, A.~X., Rish, I., and Beygelzimer, A. (2005).
\newblock Efficient test selection in active diagnosis via entropy
  approximation.
\newblock In {\em UAI}, pages 675--682.

\end{thebibliography}

\end{document}